# *Using Long Short-term Memory (LSTM) to merge precipitation data over mountainous area in Sierra Nevada*

Yihan Wang[1] Lujun Zhang[1]

1 School of Civil Engineering and Environmental Science, University of Oklahoma

**Abstract**

Obtaining reliable precipitation estimation with high resolutions in time and space is of great importance to hydrological studies. However, accurately estimating precipitation is a challenging task over high mountainous complex terrain. The three widely used precipitation measurement approaches, namely rainfall gauge, precipitation radars, and satellite-based precipitation sensors, have their own pros and cons in producing reliable precipitation products over complex areas. One way to decrease the detection error probability and improve data reliability is precipitation data merging. With the rapid advancements in computational capabilities and the escalating volume and diversity of earth observational data, Deep Learning (DL) models have gained considerable attention in geoscience. In this study, a deep learning technique, namely Long Short-term Memory (LSTM), was employed to merge a radar-based and a satellite-based Global Precipitation Measurement (GPM) precipitation product Integrated Multi-Satellite Retrievals for GPM (IMERG) precipitation product at hourly scale. The merged results are compared with the widely used reanalysis precipitation product, Multi-Radar Multi-Sensor (MRMS), and assessed against gauge observational data from the California Data Exchange Center (CDEC). The findings indicated that the LSTM-based merged precipitation notably underestimated gauge observations and, at times, failed to provide meaningful estimates, showing predominantly near-zero values. Relying solely on individual Quantitative Precipitation Estimates (QPEs) without additional meteorological input proved insufficient for generating reliable merged QPE. However, the merged results effectively captured the temporal trends of the observations, outperforming MRMS in this aspect. This suggested that incorporating bias correction techniques could potentially enhance the accuracy of the merged product.

**1 Introduction**

Precipitation is a crucial hydrological phenomenon in nature with high spatial and temporal variations. It plays a significant role in the water cycle and has essential meaning to creatures in the biosphere. Also, precipitation is usually one of the forcing parameters of hydro-meteorological models (Niu et al., 2011). Therefore, obtaining reliable precipitation observations or estimation with high resolutions in time and space is of great importance to hydrological studies and socio-economic growth.

Generally, there are three widely used precipitation measurement approaches: rainfall gauges, precipitation radars, and satellite sensors. Rainfall gauge is the most direct tool to obtain accurate rainfall observations at specific sites. However, since the distribution of rainfall gauges is uneven and sparse in most areas globally, precipitation observations from rainfall gauges are spatially discontinuous. Also, the high costs of rainfall stations limit their growth in numbers. Therefore, alternatives are needed to acquire precipitation measurements to get over the obvious disadvantage of rainfall gauges. On the other hand, precipitation radars are potent tools for obtaining real-time and high spatial resolution precipitation for vast areas. However, in complex topography, radars suffer from ground clutter and beam blockage, resulting in relative low measurement accuracy and coverage (Bartsotas et al., 2018). Satellite-based precipitation sensors compensate for the paucity of spatial coverage of radars, yet they sacrifice the spatial resolution and are also susceptible to significant uncertainties depending on the retrieval algorithms (Derin et al., 2016).

Data merging techniques have been extensively employed to combine multiple sources to decrease the detection error probability and improve data reliability (Castanedo, 2013). To obtain more reliable precipitation estimates, merging multi-source precipitation products has been extensively explored and proven to be effective (Chen et al., 2020; Li and Shao, 2010; Shrestha et al., 2011; Zhang et al., 2022). There are various statistical methods used in precipitation data merging. For example, kriging, a geostatistical interpolation technique, can be applied when the main attribute of interest is sparse, while the related other attributes are ample. Jozaghi et al. (2019) proposed an adaptive conditional bias penalized cokriging (CBPCK) technique that dynamically optimizes the fusing weights with time. Verdin et al. (2016) used the ordinary kriging together with k-nearest polynomial to merge satellite-based precipitation products with gauge observations. The statistical merging methods were able to produce promising merged products in some regions, yet they require solid mathematical assumptions (Lei et al., 2022), and have difficulty to describe the complex relationship between precipitation process and environment variables (Wu et al., 2018).

Deep Learning (DL) models have gained considerable attention in geoscience with the rapid advancements in computational capabilities and the escalating volume and diversity of earth observational data (Nearing et al., 2020; Nearing et al., 2021; Reichstein et al., 2019). They are statistical models that rely on extensive data to uncover the intrinsic relationship between input features and target variables. Among the many types of DL models, Long Short-term Memory (LSTM) (Hochreiter and Schmidhuber, 1997) is specialized in time series modeling, and thus was suitable for precipitation data series modeling. LSTM, a type of Recurrent Nural Network (RNN) architecture for sequential modeling, is renowned for learning the long-term dependencies between input and output features. Several studies have applied LSTM alone or together with other deep learning methods on precipitation data merging. For example, Wu et al. (2020) used a combined convolutional neural network (CNN) and LSTM deep fusion model to merge the Tropical Rainfall Measuring Mission (TRMM) data, Gridded satellite (GridSat-B1) data, elevation data, and in-situ rain gauge observations. In their model, CNN was used to extract the spatial features while LSTM was used to learn the feature temporal dependencies. Fan et al. (2021) compared LSTM with other 3 merging methods, namely multiple linear regression (MLR), feedforward neural networks (FNN), and random forest (RF), in merging several mainstream gridded precipitation products. Their results suggested that LSTM and RF improved the detectability of merged data. However, LSTM in general underperformed simple MLR model, probably due to small sample size.

To the author's best knowledge, there are limited number of studies on solely using LSTM to merge multi-source precipitation products over complex terrain. Firstly, previous related work did not specifically address mountainous complex areas. Secondly, while prior research merged precipitation products across multiple pixels, it did not elucidate the temporal performance of merged data at individual pixels. To further investigate the feasibility of LSTM merging method and better comprehend why LSTM has high or low effectiveness in precipitation data merging, in this project, 5 stations in the well-known complex terrain in Seirra Nevada mountain range were selected as the study locations to apply LSTM to merge 2 precipitation products, namely a radar-based National Weather Service/National Centers for Environmental Prediction (NWS/NCEP) stage IV precipitation product and a satellite-based Integrated Multi-satellitE Retrievals for GPM (IMERG) precipitation product. The LSTM model was trained to target the gauge observational data from California Data Exchange Center (CDEC). In this way, the LSTM-based merged data was supposed to incorporate precipitation estimation from all radar, satellite, and ground observations. The merged results were compared with a widely used Multi-Radar Multi-Sensor (MRMS) quantitative precipitation estimation (QPE) product which also integrates data from radar, satellite, and gauges.

The remaining sections of this paper are structured as follows. Following this introduction, Section 2 described the data and study area. Section 3 introduced the LSTM model, how LSTM model was trained, and other details in the experiment design of this project. Section 4 presented the results. Finally, Section 5 contained the key conclusions and some discussions.

**2 Data and Study Area**

2.1 Data

(1) MRMS

MRMS system combines data from multiple radars, satellites, gauge station observations, upper air observations, lightning reports, and numerical weather prediction models to generate a suite of products including QPE and flood forecasts at high spatial-temporal resolution. Hourly MRMS QPE Pass 2 products were used in this project as a comparison of merged data. They are seamless 1km mosaics that have a latency of 2 hours and include 99% of gauges. MRMS QPE was accessed at https://mtarchive.geol.iastate.edu/2022/01/31/mrms/ncep/MultiSensor_QPE_01H_Pass2/

(2) IMERG

Global Precipitation Measurement (GPM) IMERG algorithm combines information from the GPM satellite constellation to estimate precipitation over the majority of Earth's surface. GPM IMERG is a satellite-based precipitation product that provides quasi-global (60°S-60°N) precipitation estimations starting from March 2014 with three near-real-time (NRT) products, i.e., IMERG-Early, -Late, and -Final, in terms of different data latency (4h, 14h, and 3.5 months after month, respectively) and accuracy. The spatial resolution is 0.1°×0.1° (~10km at the equator). In this project, IMERG-Early (hereinafter referred to as IMERG-E) product at half hourly temporal resolution was used and downloaded from

https://disc.gsfc.nasa.gov/datasets/GPM_3IMERGHHE_06/summary?keywords=%22IMERG%20Early%22.

(3) NWS/NCEP Stage IV

The Stage IV data is a high-resolution (4km) national product based on the multi-sensor (radar and gauges) hourly/6-hourly stage III data. It is produced by 12 River Forecast Centers in the contiguous United States (CONUS). Compared to the Stage III data, Stage IV data has been operated under manual quality control by RFCs. NCEP operates mosaicking within one hour of receiving new data from RFCs. In this project, the hourly Stage IV data was downloaded from https://mesonet.agron.iastate.edu/archive/data/2021/12/01/stage4/.

(4) In-situ gauge observations

The gauge observations were used as reference and "ground truth" in this project. In-situ gauge precipitation data was obtained from the California Data Exchange Center (CDEC) managed by the California Department of Water Resources. In this project, hourly gauge observations were downloaded from https://cdec.water.ca.gov/dynamicapp/wsSensorData.

2.2 Study area

The 5 CDEC gauge stations are situated in the western United States along the Sierra Nevada (SN) mountain ranges, as illustrated in Figure 1 (left). SN is an extremely challenging region to make accurate QPE due to its complex topography and meteorological conditions. The significant mountain ranges cause localized patterns and high precipitation values that are difficult to predict. Moreover, SN receives substantial snowfall during winter months, and estimating precipitation in the form of snow and ice is a tough task.

Table 1 provides the geographical characteristics of each gauge station. Notably, ATW and MNH are near each other, both situated in the southern SN region; however, ATW boasts a slightly higher elevation. ANT and GOL are positioned in the northern SN area. In contrast, WST, located to the east of the California central valley, exhibits a relatively lower elevation compared to the other 4 stations.

The reason to pick these 5 gauge stations was that they have complete hourly precipitation records during the study period from 12/1/2021 1:00 to 2/1/2022 0:00 (1535 hours). This specific study period was chosen due to the occurrence of multiple observed precipitation events during this timeframe, as shown in Figure 1 (right). Station MNH and GOL experienced peak precipitation values, whereas moderate precipitation occurred at the rest 3 stations.

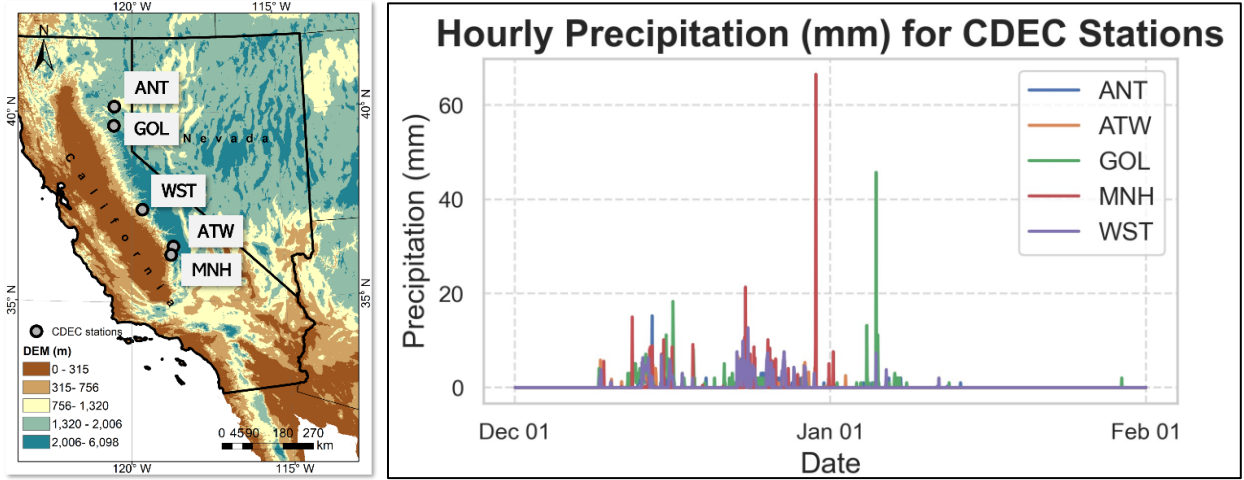

Figure 1. Study area map (left) and hourly precipitation (mm) time series during the study period (12/1/2021 1:00 – 2/1/2022 0:00) for the 5 study CDEC stations (right).

Table 1. A summary of selected CDEC gauge stations.

| Station ID | Station Name | Elevation (m) | Latitude | Longitude | Nearby City |
|---|---|---|---|---|---|
| ANT | ANTELOPE LAKE | 4960 | 40.18 | -120.61 | SUSANVILLE |
| ATW | ATWELL CAMP | 6400 | 36.46 | -118.63 | SILVER CITY |
| GOL | GOLD LAKE | 6750 | 39.67 | -120.62 | SIERRA CITY |
| MNH | MOUNTAIN HOME | 5400 | 36.24 | -118.71 | MILO |
| WST | WESTFALL | 4880 | 37.44 | -119.65 | FISH CAMP |

## 3 Methodology

### 3.1 Long Short-term Memory (LSTM)

LSTM is a type of recurrent neural network (RNN) for sequential modeling. LSTM is known for mitigating the gradient vanishing issue in conventional RNN. It introduces the gating mechanism into conventional RNN models to control the information flow. There are three gates in a LSTM model, namely input gate, forget gate, and output gate. Input gate is to decide which information to store and input into the cell state. Forget gate is to decide which information to discard from the cell state. Finally, the output gate controls what information to export from the cell state to the hidden state. The formulation of a LSTM model is listed in equation (1) to (6).

$$f_t = \sigma(W_f[X_t, h_{t-1}] + b_f) \tag{1}$$

$$i_t = \sigma(W_i[X_t, h_{t-1}] + b_i) \tag{2}$$

$$\tilde{c}_t = tanh(W_c[X_t, h_{t-1}] + b_c) \tag{3}$$

$$o_t = \sigma(W_o[X_t, h_{t-1}] + b_o) \quad (4)$$

$$c_t = f_t \cdot c_{t-1} + i_t \cdot \tilde{c}_t \quad (5)$$

$$h_t = o_t \cdot \tanh(c_t) \quad (6)$$

In equation (1) to (6), $f_t$, $i_t$, and $o_t$ represent the forget gate, input gate, and output gate, respectively. $\tilde{c}_t$ is the candidate cell state that will be used in updating new cell state $c_t$ at current time $t$. $X_t$ is the input variable. $h_{t-1}$ is the hidden state at previous time $t-1$. $W_*$ and $b_*$ (* = $f$, $i$, or $o$) are the learnable weights and bias to map $[X_t, h_{t-1}]$ to forget, input, and output gate. $\sigma$ is the sigmoid activation function ranging from 0 to 1. tanh is the tangent activation function ranging from -1 to 1.

3.2 Evaluation metrics

5 evaluation metrics were selected to assess and compare the MRMS, IMERG-E, Stage IV, and merged precipitation. The formula and description of each evaluation metric is listed in Table 2. $P_i$ and $O_i$ represented the i$^{th}$ data sample of precipitation products (i.e., MRMS, IMERG-E, Stage IV, and merged data) and observations (i.e., CDEC gauge observations), respectively. $\bar{P}$ and $\bar{O}$ represented the n sample mean. $H$ represented the number of data samples in observations that were detected in precipitation products. $M$ was the number of data samples in observations that were missed in precipitation products. CC described the linear correlation between $P$ and $O$. RMSE measured the absolute error between $P$ and $O$. RB showed the relative deviation of $P$ from $O$. POD and FAR, the categorical metrics, measured how many precipitation events were detected or missed by the precipitation products.

**Table 2. Description of evaluation metrics**

| Metric | Formula | Optimal |
|---|---|---|
| Correlation coefficient (CC) | $\dfrac{\sum_{i=1}^{n}(P_i - \bar{P})(O_i - \bar{O})}{\sqrt{\sum_{i=1}^{n}(P_i - \bar{P})^2 \sum_{i=1}^{n}(O_i - \bar{O})^2}}$ | 1 |
| Root mean square error (RMSE) | $\sqrt{\dfrac{1}{n}\sum_{i=1}^{n}(P_i - O_i)^2}$ | 1 |
| Relative bias (RB) (%) | $\dfrac{\sum_{i=1}^{n}(P_i - O_i)}{\sum_{i=1}^{n} O_i} \times 100$ | 0 |
| Probability of detection (POD) | $\dfrac{H}{H + M}$ | 1 |
| False alarm ratio (FAR) | $\dfrac{M}{H + M}$ | 0 |

3.3 Experiment design

The data required preprocessing. Specifically, IMERG-E, Stage IV, and MRMS data at the 5 gauge locations were extracted. Then, IMERG-E half hourly data was aggregated to hourly time step. In addition, any missing records in IMERG-E, Stage IV, and MRMS were linear interpolated.

The processed IMERG-E and Stage IV data were used as 2 features to train a LSTM model. The model structure and hyperparameters were manually tuned. Since there were only 2 input time series, the LSTM model had a simple structure with 1 hidden layer and 12 hidden nodes. The sequence length was 12. The learning rate was 0.001. The number of epochs was 100. In order to test the model robustness and prevent overfitting, 3-fold cross validation (CV) was employed as shown in Figure 2 (right). In this project, the term "calibration" was used in equivalent to "training" in the deep learning field, while the term "validation" is used in equivalent to "testing" in the deep learning field. Due to the presence of strong and irregular intermittent patterns in the precipitation time series, the application of 3-fold CV could evaluate the LSTM model's generalization across various temporal patterns in both calibration and validation series.

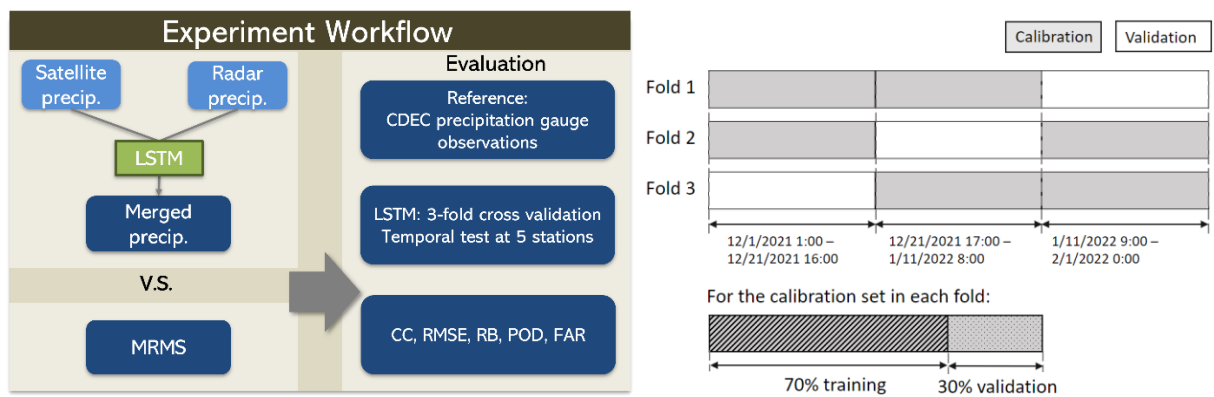

**Figure 2. An illustration of the experiment workflow (left) and 3-fold cross validation used in LSTM model calibration and validation (right). The entire time period (12/1/2021 1:00 – 2/1/2022 0:00) is evenly divided into 3 subset which forms 3 folds. For the calibration set in each fold, it is further divided into training and validation set by 70%/30%.**

## 4 Results

3.1 Time series comparison for each fold's calibration and validation results

Figure 3 depicted the time series at 5 CDEC gauge station locations for 3 folds. The performance of merged precipitation data was different given different calibration and validation sets. For ANT, fold-1 showed very unfavorable merged results which were basically all 0 values. However, for fold-2 and fold-3, ANT produced better merged results as they were able to at least capture some precipitation that occurred from 2021/12 to 2022/1. For ATW, fold-3 showed some precipitation, nevertheless, the merged precipitation was very misinformative as it was almost unable to suggest any precipitation. Similarly, merged precipitation at MNH was also unplausible compared to any other precipitation products. However, for GOL and WST, merged precipitation could produce precipitation for all folds, despite the fact that the magnitude of CDEC gauge precipitation was underestimated by merged precipitation very significantly.

Figure 4 showed the stitched time series at 5 CDEC station locations from the testing period for each fold. Consistent with the time series in Figure 3, station ATW and MNH still had the worst merged precipitation compared to any other precipitation products, where basically the merged precipitation was always 0. At ATW, despite the peak precipitation values from both IMERG-E and Stage IV, the merged data was unable to learn such information from either source. On the contrary, at ANT, GOL, and WST, merged precipitation exhibited better performance in capturing the temporal trend of gauge observations. At GOL, the peak precipitation from gauge observations at around $800^{th}$ hour was missed by all individual products. Therefore, it was not surprising that merged data missed that, too. At WST, Stage IV significantly overestimated gauge data, especially from $500^{th}$ to $700^{th}$ hour. However, merged data was able to correct these overestimations, although the correction was too much which finally led to underestimation.

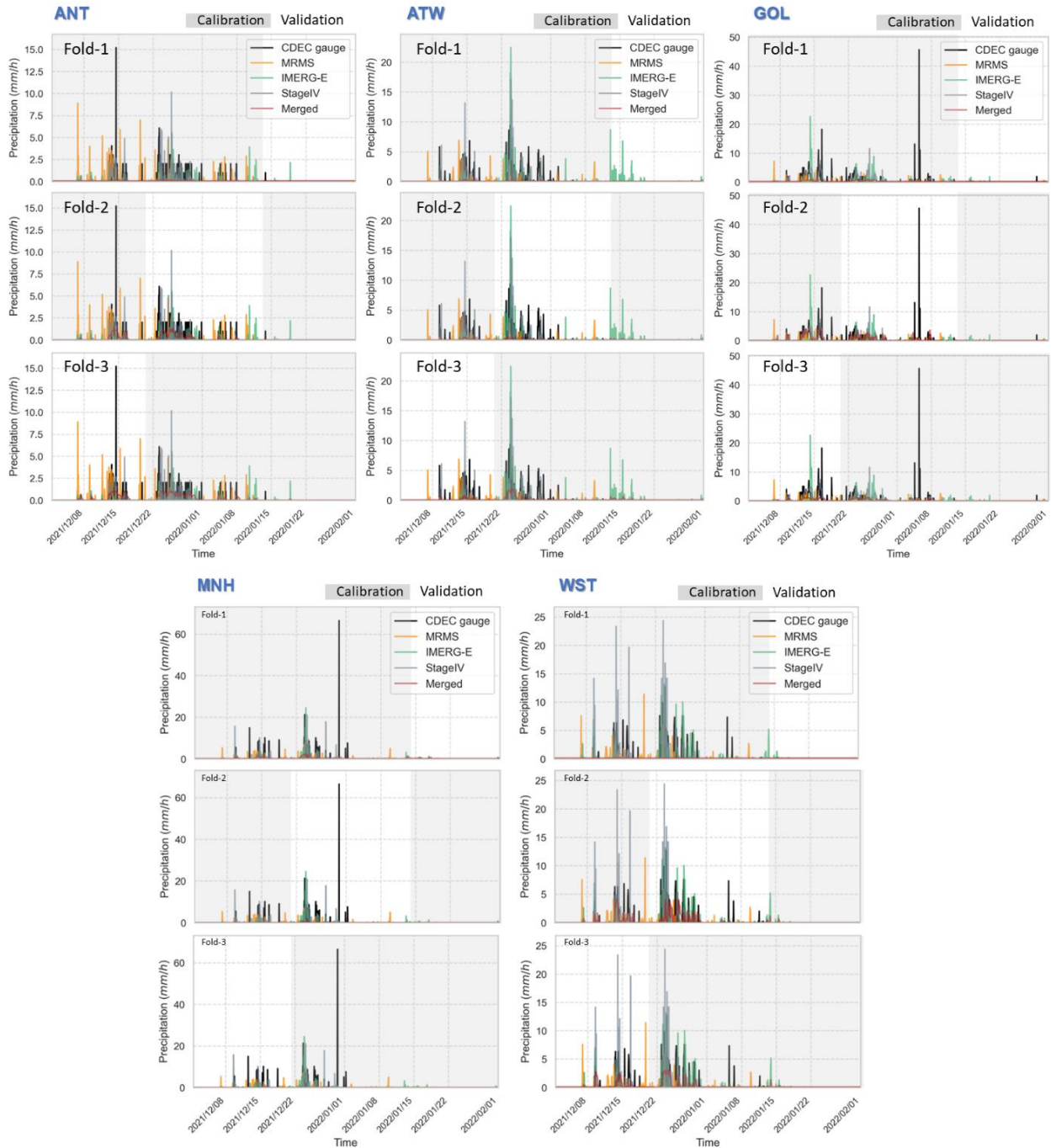

**Figure 3. Time series for the entire calibration and validation periods of 3-fold cross validation. Shadowed area represents calibration period whereas white background represents validation period for each fold.**

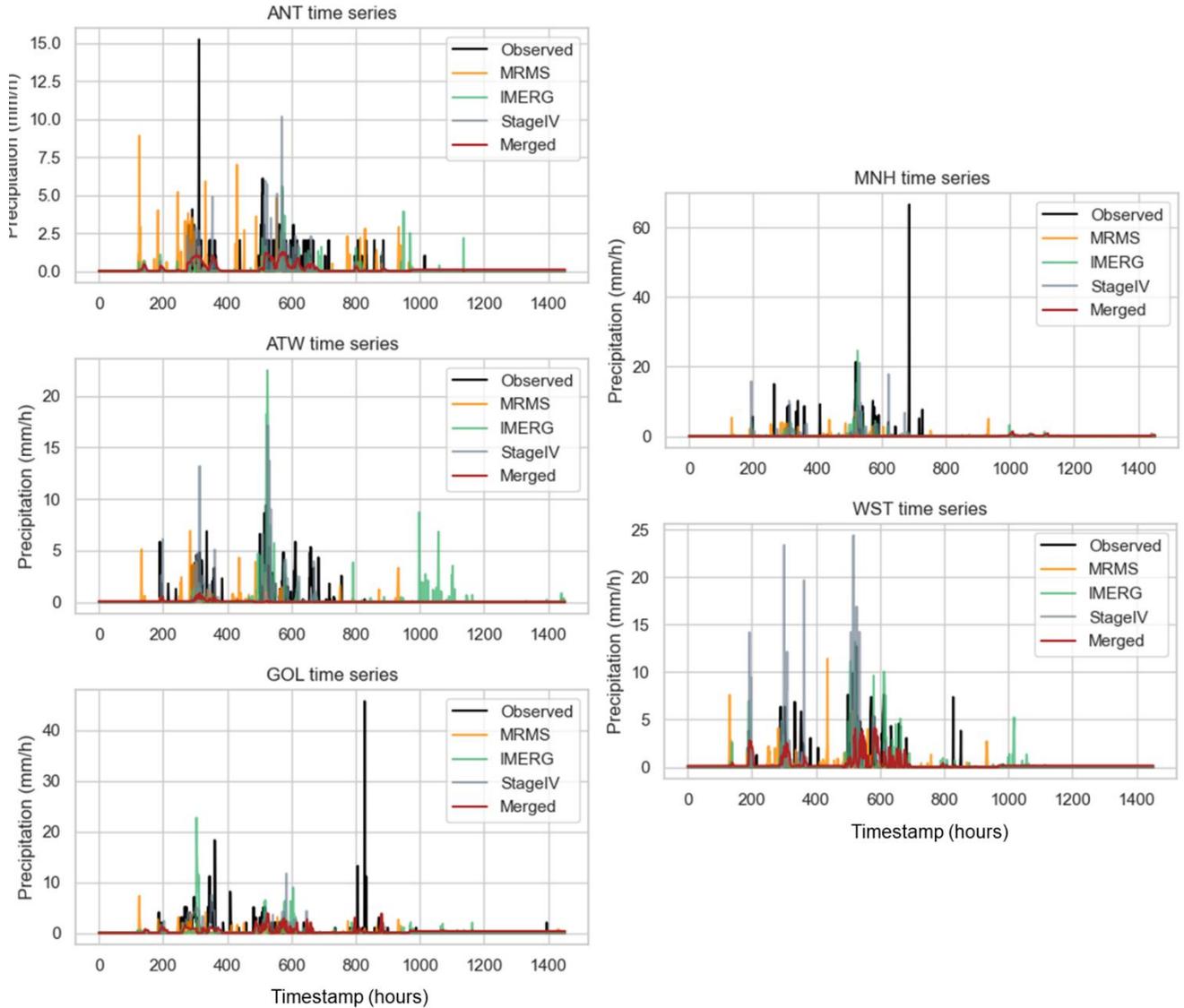

**Figure 4. Time series for the stitched validation periods from each fold in 3-fold cross validation.**

3.2 Scatter plot comparison for each fold's validation results

In order to gain a further understanding of the underestimation or overestimation patterns between each QPE and gauge observations, the scatter plot between CDEC data and MRMS, IMERG-E, Stage IV, and merged product was plotted in logarithm scale. Since there were a lot of 0 (or near 0) values, the range of the logarithm values were restricted from -2 to 2 to eliminate the clustered values at around 0. Figure 5 showed that merged precipitation had apparent underestimation for precipitation higher than 1mm/hour for all stations, which was consistent with the time series comparison in Figure 3 and 4. For small precipitation < 1mm/hour, GOL and WST showed that merged precipitation overestimated observations, whereas underestimation still exhibited at ANT, ATW, and MNH. At all 5 station locations, overestimation was observed in Stage IV and IMERG-E estimates. However, merged precipitation did not learn the high values from either of its sources at ATW or MNH.

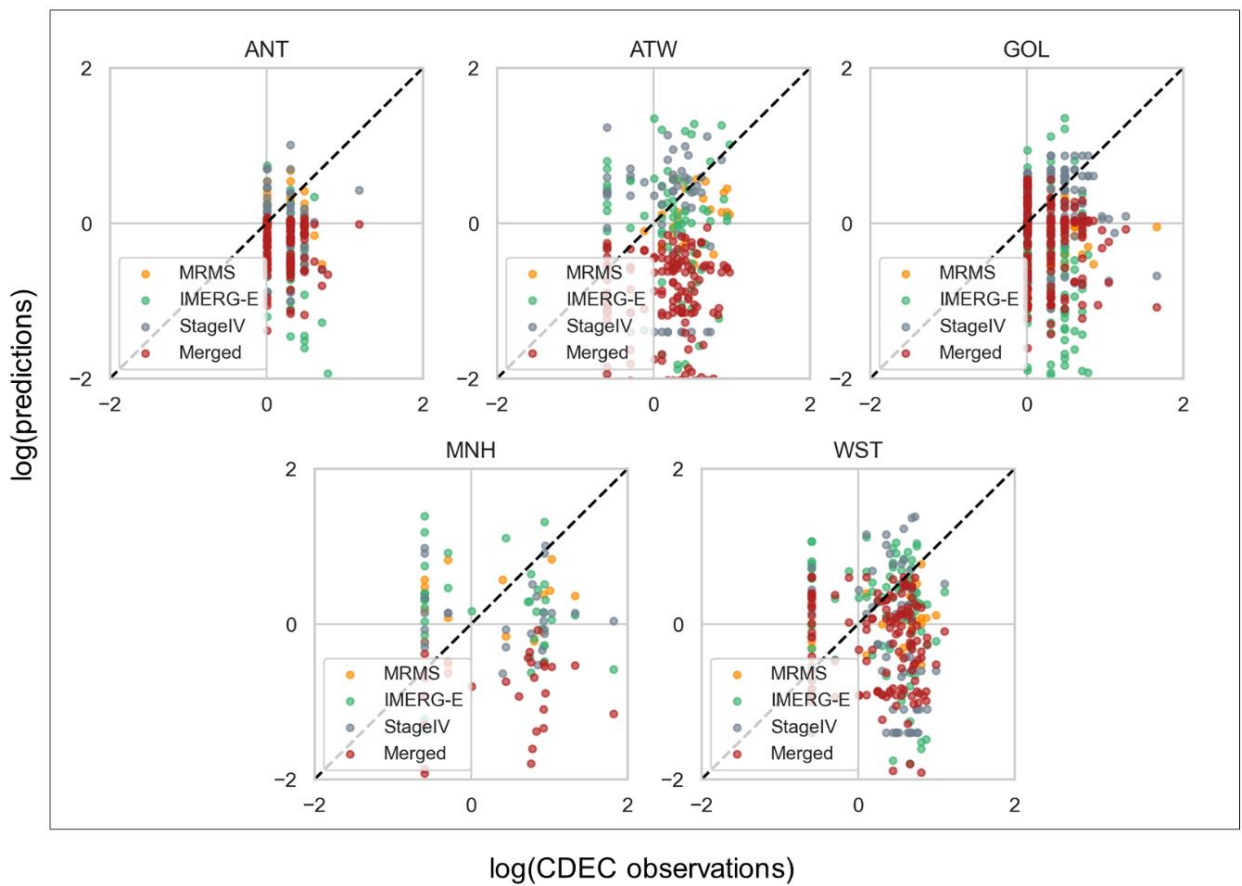

**Figure 5. Scatter plot for log(CDEC observations) versus log(predictions) from MRMS, IMERG-E, StageIV, and merged precipitation products, respectively. The range is limited to -2 to 2 so that 0 (or almost 0) precipitation values are eliminated in the plots.**

3.3 Evaluation metrics for each fold's validation results

Table 3 presented the statistics for 5 evaluation metrics for all individual and merged precipitation estimates for the stitched validation results. Despite the unfavorable performance at ANT as shown in Figure 3, 4, and 5, the merged precipitation data surprisingly produced the best CC, RMSE, RB, and POD. MRMS showed the worst CC among all products, which was consistent with the time series in Figure 4. CC describes the increasing and decreasing trend between two data series. Merged data, as shown in Figure 4, did have the most similar trend as CDEC data even though the magnitude was off. The lower RMSE of merged data was also explainable. The overestimation by the rest of the products (MRMS, IMERG, and Stage IV) was very significant. Therefore, one could expect they had higher RMSE mainly introduced by those overestimations. Regarding the POD, merged precipitation tended to always produce some value at almost all the time stamps. In other words, LSTM model tended to not produce exactly 0 values but rather some small values. This phenomenon could also explain the reason why merged data always exhibited highest FAR, as there was rarely exact 0 values in merged data.

By looking at the 2 CDEC stations where merged data basically was meaningless, i.e., ATW and MNH, the statistics of merged data were not good, but not unacceptably bad, either. Specifically, its CC at ATW ranked slightly better than MRMS's, and its POD ranked the second best one among all four products. However, for MNH the merged data was consistently the worst one. Such observations suggested that through LSTM training the merged data might be able to learn the trend of observation data. If some bias correction techniques could be applied to these badly trained results, the merged results accuracy could probably be improved.

For the relatively well-trained merged results at WST, merged data showed better CC than MRMS, probably because its sources – IMERG-E and Stage IV – had high CC values. The reason that merged data had the best RMSE, RB, and POD, as well as the worst FAR could be explained the same way as for ANT.

**Table 3. A summary of evaluation metrics (i.e., CC, RMSE, RB in %, POD, and FAR) for each CDEC station based on the stitched validation periods from each fold in 3-fold cross validation.**

ANT

| Metric | Merged | MRMS | IMERG | StageIV |
|---|---|---|---|---|
| CC | 0.383 | 0.093 | 0.112 | 0.264 |
| RMSE | 0.653 | 0.872 | 0.751 | 0.794 |
| RB (%) | -5.724 | -26.456 | -63.556 | -19.114 |
| POD | 0.940 | 0.265 | 0.419 | 0.615 |
| FAR | 0.882 | 0.737 | 0.742 | 0.673 |

ATW

| Metric | Merged | MRMS | IMERG | StageIV |
|---|---|---|---|---|
| CC | 0.229 | 0.214 | 0.304 | 0.301 |
| RMSE | 0.905 | 0.937 | 1.387 | 1.129 |
| RB (%) | -83.879 | -60.630 | 8.979 | -13.782 |
| POD | 0.444 | 0.172 | 0.523 | 0.404 |
| FAR | 0.865 | 0.729 | 0.703 | 0.579 |

GOL

| Metric | Merged | MRMS | IMERG | StageIV |
|---|---|---|---|---|
| CC | 0.149 | 0.106 | 0.061 | 0.301 |
| RMSE | 1.691 | 1.724 | 1.999 | 1.708 |
| RB (%) | -15.932 | -75.694 | -52.308 | 5.072 |
| POD | 0.754 | 0.251 | 0.404 | 0.837 |
| FAR | 0.840 | 0.605 | 0.655 | 0.514 |

MNH

| Metric | Merged | MRMS | IMERG | StageIV |
|---|---|---|---|---|
| CC | -0.047 | 0.106 | 0.094 | 0.107 |
| RMSE | 2.154 | 2.15 | 2.403 | 2.345 |
| RB (%) | -75.276 | -57.906 | -1.101 | -1.114 |
| POD | 0.044 | 0.235 | 0.544 | 0.456 |
| FAR | 0.994 | 0.84 | 0.833 | 0.807 |

WST

| Metric | Merged | MRMS | IMERG | StageIV |
|---|---|---|---|---|
| CC | 0.257 | 0.143 | 0.295 | 0.269 |
| RMSE | 1.161 | 1.223 | 1.383 | 1.684 |
| RB (%) | -3.700 | -66.967 | -19.665 | -6.899 |
| POD | 0.910 | 0.181 | 0.556 | 0.479 |
| FAR | 0.890 | 0.778 | 0.644 | 0.648 |

## 5 Discussions and Conclusions

Precipitation estimation presents significant challenges in regions with complex terrain. Traditional methods such as rainfall gauges, radar, and satellite sensors, while having their respective advantages, encounter difficulties in providing accurate estimates under such conditions. To harness the strengths of various precipitation estimation methods and mitigate individual data errors, the merging of precipitation data has been actively explored. The growing computational capacity and the abundance of Earth observational data have led to an increased interest in employing deep learning techniques for precipitation data merging. In this study, the Long Short-Term Memory (LSTM) deep learning method was employed to merge precipitation data from radar-based (Stage IV) and satellite-based (IMERG-E) Quantitative Precipitation Estimates (QPE). The efficiency of LSTM models in precipitation data merging, particularly over complex terrain, was assessed. The LSTM merged product was compared with the widely used reanalysis precipitation product MRMS, using the reference CDEC gauge stations. The study period, spanning from 12/1/2021 1:00 to 2/1/2022 0:00 at hourly time scale, covered multiple observed precipitation events. The dataset's ample size provided sufficient samples for training and testing the LSTM model.

In conclusion, the results of this project showed that:

(1) Only using precipitation estimates from Stage IV and IMERG-E was not sufficient to generate reliable and helpful merged precipitation product. The merged QPE significantly underestimated the gauge observations. More meteorological and geographical variables are needed as ancillary inputs to the LSTM model. For example, Derin and Kirstetter (2022) indicated that moisture flux convergence is a significant factor contributing to intense rainfall. Therefore, the incorporation of horizontal moisture flux convergence and vertical motions into the LSTM merging model probably could mitigate the underestimation issue in the merged QPE.
(2) Despite the unfavorable performance of LSTM-based merged data, it was able to capture the temporal trend of the gauge observations suggested by its relatively high CC values. Therefore, by incorporating some bias correction techniques perhaps the accuracy of current merged data could be improved.
(3) In the cases where LSTM-based merged results produced meaningful precipitation estimates, the merged results had higher POD and consistently higher FAR compared to other precipitation estimates from MRMS, Stage IV, and IMERG-E. The reason was the merged data constantly had some small precipitation values even when there was no precipitation in observational data. Therefore, by solely looking at POD and FAR without considering the precipitation magnitude would give biased conclusion. As suggested by Fan et al. (2021), LSTM- and RF-based merged products improved the POD. However, according to the result in this project, this conclusion probably needed further investigation.

**References**

Bartsotas, N., Anagnostou, E., Nikolopoulos, E. and Kallos, G. 2018. Investigating satellite precipitation uncertainty over complex terrain. Journal of Geophysical Research: Atmospheres 123(10), 5346-5359.

Castanedo, F. 2013. A review of data fusion techniques. The scientific world journal 2013.

Chen, F., Gao, Y., Wang, Y. and Li, X. 2020. A downscaling-merging method for high-resolution daily precipitation estimation. Journal of Hydrology 581, 124414.

Derin, Y., Anagnostou, E., Berne, A., Borga, M., Boudevillain, B., Buytaert, W., Chang, C.-H., Delrieu, G., Hong, Y. and Hsu, Y.C. 2016. Multiregional satellite precipitation products evaluation over complex terrain. Journal of Hydrometeorology 17(6), 1817-1836.

Derin, Y. and Kirstetter, P.E. 2022. Evaluation of IMERG over CONUS complex terrain using environmental variables. Geophysical Research Letters 49(19), e2022GL100186.

Fan, Z., Li, W., Jiang, Q., Sun, W., Wen, J. and Gao, J. 2021. A comparative study of four merging approaches for regional precipitation estimation. IEEE Access 9, 33625-33637.

Hochreiter, S. and Schmidhuber, J. 1997. Long short-term memory. Neural computation 9(8), 1735-1780.

Jozaghi, A., Nabatian, M., Noh, S., Seo, D.-J., Tang, L. and Zhang, J. 2019. Improving multisensor precipitation estimation via adaptive conditional bias–penalized merging of rain gauge data and remotely sensed quantitative precipitation estimates. Journal of Hydrometeorology 20(12), 2347-2365.

Lei, H., Zhao, H. and Ao, T. 2022. A two-step merging strategy for incorporating multi-source precipitation products and gauge observations using machine learning classification and regression over China. Hydrology and Earth System Sciences 26(11), 2969-2995.

Li, M. and Shao, Q. 2010. An improved statistical approach to merge satellite rainfall estimates and raingauge data. Journal of Hydrology 385(1-4), 51-64.

Nearing, G., Kratzert, F., Klotz, D., Hoedt, P.-J., Klambauer, G., Hochreiter, S., Gupta, H., Nevo, S. and Matias, Y. 2020 A deep learning architecture for conservative dynamical systems: Application to rainfall-runoff modeling.

Nearing, G.S., Kratzert, F., Sampson, A.K., Pelissier, C.S., Klotz, D., Frame, J.M., Prieto, C. and Gupta, H.V. 2021. What role does hydrological science play in the age of machine learning? Water Resources Research 57(3), e2020WR028091.

Niu, G.Y., Yang, Z.L., Mitchell, K.E., Chen, F., Ek, M.B., Barlage, M., Kumar, A., Manning, K., Niyogi, D. and Rosero, E. 2011. The community Noah land surface model with multiparameterization options (Noah-MP): 1. Model description and evaluation with local-scale measurements. Journal of Geophysical Research: Atmospheres 116(D12).

Reichstein, M., Camps-Valls, G., Stevens, B., Jung, M., Denzler, J., Carvalhais, N. and Prabhat, f. 2019. Deep learning and process understanding for data-driven Earth system science. Nature 566(7743), 195-204.

Shrestha, R., Houser, P.R. and Anantharaj, V.G. 2011. An optimal merging technique for high-resolution precipitation products. Journal of Advances in Modeling Earth Systems 3(4).

Verdin, A., Funk, C., Rajagopalan, B. and Kleiber, W. 2016. Kriging and local polynomial methods for blending satellite-derived and gauge precipitation estimates to support hydrologic early warning systems. IEEE Transactions on Geoscience and Remote Sensing 54(5), 2552-2562.

Wu, H., Yang, Q., Liu, J. and Wang, G. 2020. A spatiotemporal deep fusion model for merging satellite and gauge precipitation in China. Journal of Hydrology 584, 124664.

Wu, Z., Zhang, Y., Sun, Z., Lin, Q. and He, H. 2018. Improvement of a combination of TMPA (or IMERG) and ground-based precipitation and application to a typical region of the East China Plain. Science of the Total Environment 640, 1165-1175.

Zhang, J., Xu, J., Dai, X., Ruan, H., Liu, X. and Jing, W. 2022. Multi-source precipitation data merging for heavy rainfall events based on cokriging and machine learning methods. Remote Sensing 14(7), 1750.